\documentclass[10pt,twocolumn,letterpaper]{article}

\usepackage{cvpr}
\usepackage{times}

\usepackage{graphicx}
\graphicspath{{figures/}}

\usepackage{amsmath}
\usepackage{amssymb}
\usepackage{tabularx}
\usepackage[utf8]{inputenc}
\usepackage[T1]{fontenc}
\usepackage{textcomp}
\usepackage{gensymb}
\usepackage{multirow}
\usepackage{capt-of}
\usepackage{etoolbox}
\usepackage{adjustbox}

\usepackage{float}
\usepackage{algorithm}
\usepackage[noend]{algpseudocode}
\newfloat{algorithm}{t}{lop}

\usepackage{authblk}

\usepackage{cite}

\usepackage{multicol}

\usepackage[usenames,dvipsnames]{xcolor}

\newcommand{\gradslam}{$\nabla$SLAM}
\newcommand{\gradlm}{$\nabla$LM}

\newcommand{\mb}[1]{\mathbf{#1}}

\usepackage[breaklinks=true,bookmarks=false,pagebackref,colorlinks=true]{hyperref}

\cvprfinalcopy % *** Uncomment this line for the final submission

\def\cvprPaperID{****} % *** Enter the CVPR Paper ID here

\begin{document}

\makeatletter
\renewcommand\AB@affilsepx{, \protect\Affilfont}
\makeatother

%%%%%%%%% TITLE
\title{\gradslam: Automagically differentiable SLAM \\ \large 
{\normalfont  \textbf{\url{https://gradslam.github.io}}} \vspace{-1cm}}

% \author{
% Krishna Murthy J.$^{1,2}$\thanks{Correspondence to krrish94 [at] gmail [dot] com}\\
% $^{1}$Universit\'e de Montr\'eal
% \and
% Ganesh Iyer$^{3}$\\
% $^{2}$Mila
% \and
% Liam Paull$^{1,2}$\\
% $^{3}$Carnegie Mellon University
% }

\author[1,2,3]{Krishna Murthy J.\thanks{Equal contribution}}
\author[4]{Soroush Saryazdi$^*$}
\author[5]{Ganesh Iyer}
\author[1,2,3,6]{Liam Paull\thanks{No $\nabla$ students were harmed in the making of this work.}}
\affil[1]{Universit\'e de Montr\'eal}
\affil[2]{Mila}
\affil[3]{Robotics and Embodied AI Lab (\href{http://montrealrobotics.ca}{REAL})}
\affil[4]{Concordia University}
\affil[5]{Carnegie Mellon University}
\affil[6]{Candian CIFAR AI Chair}

%%%%%%%%%%%%%%%%%%%%%%%%%%
% Teaser figure
\makeatletter
\let\@oldmaketitle\@maketitle
\renewcommand{\@maketitle}{\@oldmaketitle
\centering
\vspace{-1cm}
\includegraphics[width=\textwidth]{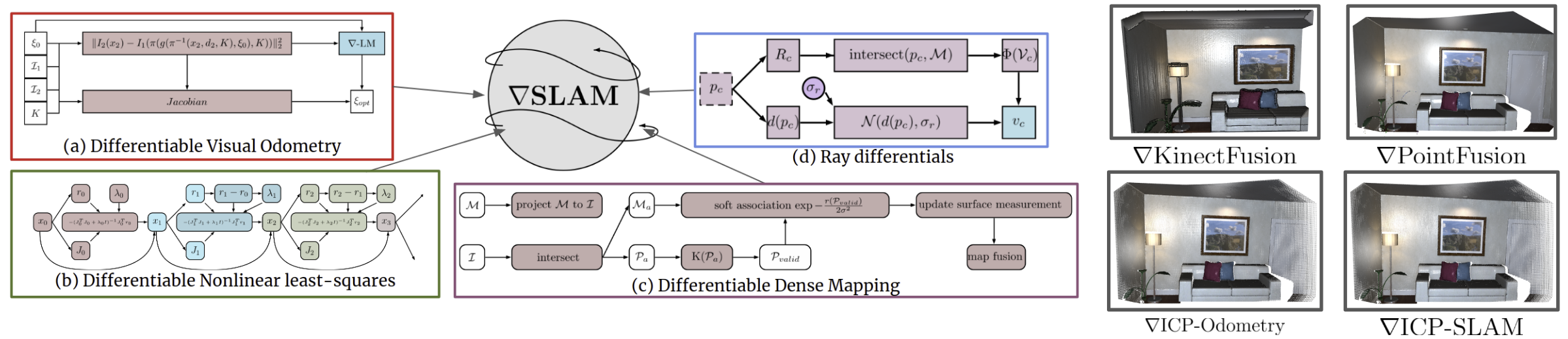}
\captionof{figure}{
\textbf{\gradslam{} (gradSLAM)} is a \emph{fully differentiable} dense simultaneous localization and mapping (SLAM) system. The central idea of \gradslam{} is to construct a computational graph representing every operation in a dense SLAM system. We propose differentiable alternatives to several non-differentiable components of traditional dense SLAM systems, such as optimization, odometry estimation, raycasting, and map fusion. This creates a pathway for gradient-flow from 3D map elements to sensor observations (e.g., \emph{pixels}). We implement differentiable variants of three dense SLAM systems that operate on voxels, surfels, and pointclouds respectively. \gradslam{} thus is a novel paradigm to integrate representation learning approaches with classical SLAM.
}
\label{fig:teaser_new}
\vspace{0.3cm}
}
\makeatother
% End teaser figure
%%%%%%%%%%%%%%%%%%%%%%%%%%

\maketitle
%\thispagestyle{empty}

%%%%%%%%% ABSTRACT
\begin{abstract}
\vspace{-0.3cm}
% The question of ``representation" is central in the context of dense simultaneous localization and mapping (SLAM). Learning-based approaches have the potential to leverage data or task performance to directly inform the representation.
% However,
Blending representation learning approaches with simultaneous localization and mapping (SLAM) systems is an open question, because of their highly modular and complex nature.
Functionally, SLAM is an operation that transforms raw sensor inputs into a distribution over the state(s) of the robot and the environment. If this \emph{transformation} (SLAM) were expressible as a differentiable function, we could leverage task-based error signals to learn representations that optimize task performance.
However, several components of a typical dense SLAM system are non-differentiable.
In this work, we propose \textbf{\gradslam{} (\emph{grad}SLAM)}, a methodology for posing SLAM systems as \emph{differentiable} computational graphs, which unifies gradient-based learning and SLAM.
We propose differentiable trust-region optimizers, surface measurement and fusion schemes, and raycasting, without sacrificing accuracy.
This amalgamation of dense SLAM with computational graphs enables us to backprop all the way from 3D maps to 2D pixels, opening up new possibilities in gradient-based learning for SLAM\footnote{Video abstract: \href{https://youtu.be/2ygtSJTmo08}{https://youtu.be/2ygtSJTmo08}}.

\textbf{TL;DR}: We leverage the power of automatic differentiation frameworks to make dense SLAM differentiable.
\end{abstract}

\section{Introduction}
\label{sec:intro}

Simultaneous localization and mapping (SLAM) has---for decades---been a central problem in robot perception and state estimation.
% is a key enabler for robot systems operating in previously unknown environments. 
A large portion of the SLAM literature has focused either directly or indirectly on the question of map representation. This fundamental choice dramatically impacts the choice of processing blocks in the SLAM pipeline, as well as all other downstream tasks that depend on the outpus of the SLAM system. 
% such as data association and optimization. 
Of late, gradient-based learning approaches have transformed the outlook of several domains (Eg. image recognition \cite{alexnet}, language modeling \cite{transformer}, speech recognition \cite{hinton2012deep}). However, such techniques have had limited success in the context of SLAM, primarily since many of the elements in the standard SLAM pipeline are not differentiable. A fully differentiable SLAM system would enable task-driven representation learning since the error signals indicating task performance could be back-propagated all the way through the SLAM system, to the raw sensor observations.

This is particularly true for \emph{dense 3D maps} generated from RGB-D cameras, where there has been a lack of consensus on the right representation (pointclouds, meshes, surfels, etc.). 
%Creating rich 3D representations of a scene from an image has for long been an exciting area of research. 
%Consequently, several systems to build such 3D scene representations (aka maps) from a sequence of images have been proposed (cite a few
Several methods have demonstrated a capability for producing dense 3D maps from sequences of RGB or RGB-D frames \cite{kinectfusion, kintinuous,pointfusion}.
%SLAM / RGB-D reconstruction papers). 
However, none of these methods are able to solve the \textit{inverse mapping} problem, i.e., answer the question: ``\emph{How much does a specific pixel-measurement contribute to the resulting 3D map}"? Formally, we desire an the expression that relates a pixel in an image (or in general, a sensor measurement \emph{s}) to a 3D map $\mathcal{M}$ of the environment. 
%It's usually very hard to come up with a reasonable answer to this question. 
We propose to solve this through the development of a differentiable mapping function $\mathcal{M} =  \mathcal{G}_{SLAM}(s)$.  Then the gradient of that mapping $\nabla_s \mathcal{M}$ can intuitively tell us that \emph{perturbing the sensor measurement $s$ by an infinitesimal $\delta s$ causes the map $\mathcal{M}$ to change by $\nabla_s \mathcal{G}_{SLAM}(s) \delta s$}. 
%But, such a smooth mapping $g_{SLAM}$ does not exist. \stream{We set out to do that, with \gradslam{}. A very rough sketch of the motivation.}

Central to our goal of realizing a fully differentiable SLAM system are \emph{computational graphs}, which underlie most gradient-based learning techniques. We make the observations that, if an entire SLAM system can be decomposed into elementary operations, all of which are differentiable, we could compose these elementary operations\footnote{Again, using differentiable composition operators.} to preserve differentiability. However, modern \emph{dense} SLAM systems are quite sophisticated, with several non-differentiable subsystems (optimizers, raycasting, surface mapping), that make such a construct challenging.

We propose \gradslam{} (\emph{grad}SLAM), a differentiable computational graph view of SLAM. We show how \emph{all} non-differentiable functions in SLAM can be realised as smooth mappings. First, we propose a differentiable trust region optimizer for nonlinear least squares systems. Building on it, we present differentiable strategies of mapping, raycasting, and global measurement fusion.

% differtiable versions of nonlinear least squares optimization, raycasting, and rasterization differentiable, without sacrificing a great deal in terms of performance or accuracy. 

% To the best of our knowledge, we are the first to introduce a differentiable parameterization of dense mapping. While existing approaches use neural networks as \emph{differentiable maps} (eg. for relocalization \cite{posenet}), the map is represented \emph{implicitly} (i.e., encoded in the weights of a neural network). We conversely demonstrate how the dense mapping process can be approximated as a differentiable operation, and present a wholly differentiable SLAM system.

The \gradslam{} framework is very general, and can be extended most dense SLAM systems for differentiability. In Sec.~\ref{sec:case_studies}, we provide three examples of SLAM systems that can be realized as differentiable computation graphs: implicit-surface mapping (Kinectfusion \cite{kinectfusion}), surfel-based mapping (PointFusion \cite{pointfusion}), and iterative closest point (ICP) mapping (ICP-SLAM). We show that the differentiable approaches maintain similar performance to their non-differentiable counterparts, with the added advantage that they allow gradients to flow through them.

To foster further research on differentiable SLAM systems and their applications to spatially-grounded learning, \gradslam{} is available as an open-source PyTorch framework. Our project page and code can be accessed at \href{https://gradslam.github.io}{https://gradslam.github.io}.

\section{Related Work}
\label{sec:relatedwork}

Several works in recent years have applied recent machine learning advances to SLAM or have reformulated a subset of \emph{components} of the full SLAM system in a differentiable manner. 
% We provide a (non-exhaustive) overview of a few attempts at infusing differentiability into components of SLAM systems.

%In this section, we present a brief recap of approaches that leverage differentiability of sub-components of a SLAM system.

\subsection{\textbf{Learning-based SLAM approaches}}

There is a large body of work in deep learning-based SLAM systems. For example, CodeSLAM \cite{codeslam} and SceneCode \cite{scenecode} attempt to represent scenes using compact \emph{codes} that represent. 2.5D depth map. 
%At inference time, these latent codes are optimized over, as variables in a pose graph. BA-Net \cite{banet} uses a set of basis depth maps which can be fused to predict 2.5D depths and features a learned optimizer. 
DeepTAM \cite{deeptam} trains a tracking network and a mapping network, which learn to reconstruct a voxel representation from a pair of images.
CNN-SLAM \cite{cnnslam} extends LSD-SLAM \cite{lsdslam}, a popular monocular SLAM system, to use single-image depth predictions from a convnet.
Another recent trend has been to try to formulate the SLAM problem over higher level features such as objects, which may be detected with learned detectors \cite{CubeSLAM}\cite{Mu_IROS_2016}\cite{Parkhiya_ICRA_2018}.
DeBrandandere et al. \cite{DeBradandere} perform lane detection by backpropagating least squares residuals into a frontend module. Recent work has also formulated the passive \cite{posenet} and active localization problems \cite{anl, dal} in an end-to-end differentiable manner.
While all of these approaches try to leverage differentiability in submodules of SLAM systems (eg. odometry, optimization, etc.), there is no single framework that models an entire SLAM pipeline as a differentiable graph.

\subsection{\textbf{Differentiable visual odometry}}

The beginnings of differentiable visual odometry can be traced back to the seminal Lucas-Kanade iterative matching algorithm \cite{lucas1981iterative}. Kerl \etal{} \cite{kerl2013vo}\footnote{The formulation first appeared in Steinbr\"uker \etal{} \cite{steinbrucker2011real}.} apply the Lucas-Kanade algorithm to perform real-time dense visual odometry. Their system is differentiable, and has been extensively used for self-supervised depth and motion estimation \cite{garg2016unsupervised, sfmlearner, undeepvo}. Coupled with the success of Spatial Transformer Netowrks (STNs) \cite{stn}, several libraries (gvnn \cite{gvnn}, kornia \cite{kornia}) have since implemented these techniques as differentiable \emph{layers}, for use in neural networks.

However, extending differentiability beyond the two-view case (\emph{frame-frame alignment}) is not straightforward. Global consistency necessitates fusing measurements from live frames into a global model (\emph{model-frame alignment}), which is not trivially differentiable.

\subsection{\textbf{Differentiable optimization}}

Some approaches have recently proposed to learn the optimization of nonlinear least squares objective functions. This is motivated by the fact that similar cost functions have similar loss landscapes, and learning methods can help converge faster, or potentially to better minima.

In BA-Net \cite{banet}, the authors learn to predict the damping coefficient of the Levenberg-Marquardt optimizer, while in LS-Net \cite{lsnet}, the authors entirely replace the Levenberg-Marquard optimizer by an LSTM netowrk \cite{LSTM} that predicts update steps.
In GN-Net \cite{GNNet}, a differentiable version of the Gauss-Newton loss is used to show better robustness to weather conditions. RegNet \cite{RegNET} employs a learning-based optimization approach based on photometric error for image-to-image pose registration. However, all the aforementioned approaches require the training of additional neural nets and this requirement imposes severe limitations on the generalizability. OptNet \cite{optnet} introduces differentiable optimization layers for quadratic programs, that do not involve learnable parameters.
%https://arxiv.org/abs/1902.00293 observes that we can backprop through a least squares module. Moved to end-to-end section

Concurrently, Grefenstette \etal{}~\cite{higher} propose to unroll optimizers as computational graphs, which allows for computation of arbitrarily higher order gradients. Our proposed differentiable Levenberg-Marquardt optimizer is similar in spirit, with the addition of gating functions to result in better gradient flows.

%\stream{Need to add \emph{GN-Net: The Gauss-Newton loss for multi-weather relocalization}, \emph{RegNet: Learning the Optimization of% Direct Image-to-Image Pose Registration}}

% \TODO{Cite "OptNet: Differentiable Optimization as a Layer in Neural Networks".} This paper talks about formulating quadratic progams as differentiable layers. In \gradslam{}, we are concerned with nonlinear least squares.

In summary, to the best of our knowledge, there is no \emph{single} approach that models the entire SLAM pipeline as a differentiable model, and it is this motivation that underlies \gradslam{}.

% \LP{Need summarizing statement about the approach of grad slam}

\section{\gradslam{}}
\label{sec:gradslam}

In this section we will overview our proposed method for \gradslam{} and also detail the individual differentiable sub-components. 

\iffalse
\begin{figure*}
    \centering
    \includegraphics[scale=0.1]{figures/pipeline_idea.jpg}
    \caption{Caption}
    \label{fig:pipeline_idea}
\end{figure*}
\fi
\newcommand{\forward}{\texttt{forward}}
\newcommand{\gradparams}{\texttt{grad\_parameters}}
\newcommand{\gradinput}{\texttt{grad\_input}}

\subsection{\textbf{Preliminaries: Computational graphs}}

\begin{figure}[!h]
    \centering
    \includegraphics[width=0.9\columnwidth]{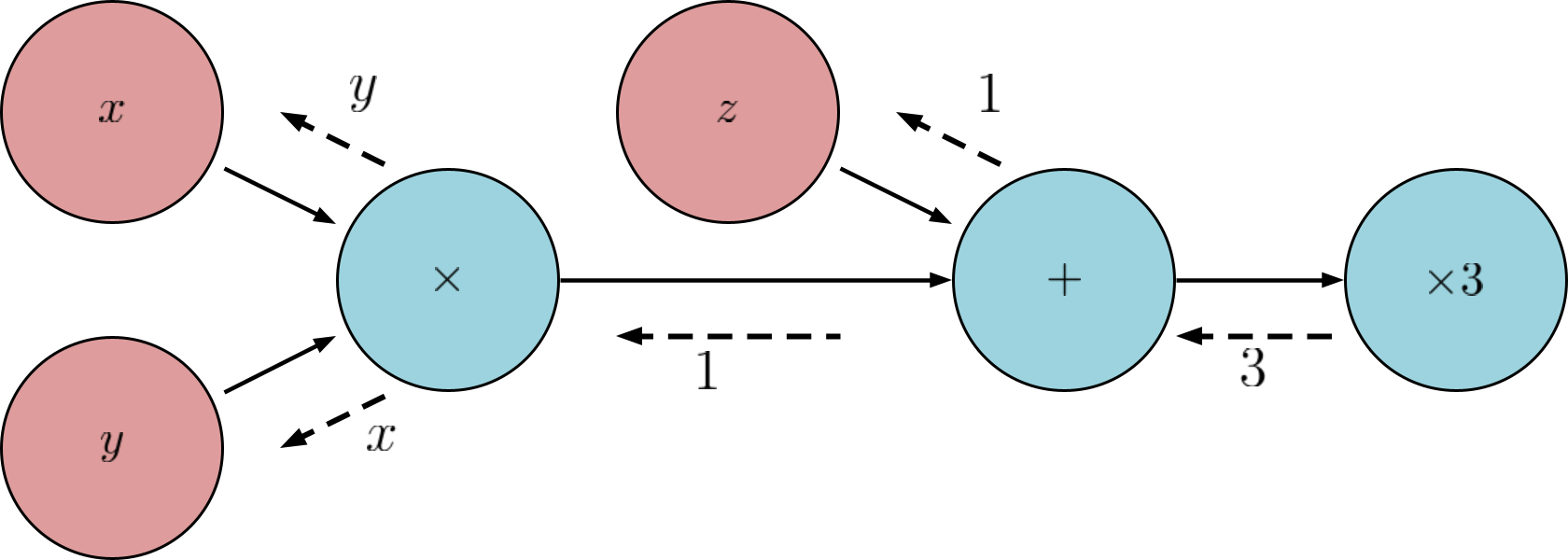}
    \caption{\textbf{A computational graph}. Nodes in {\textcolor{red}{red}} represent variables. Nodes in {\textcolor{blue}{blue}} represent operations on variables. Edges represent data flow. This graph computes the function $3(xy+z)$. Dashed lines indicate (local, i.e., per-node) gradients in the backward pass.}
    \label{fig:computational_graph_mockup}
\end{figure}

%\TODO{Talk about probabilistic programming? (somewhat related)}

In gradient-based learning architectures, all functions and approximators are conventionally represented as \emph{computational graphs}. Formally, a computation graph is a directed acyclic graph $\mathcal{G} = \left(\mathcal{V}, \mathcal{E}\right)$, where each node $v \in \mathcal{V}$ holds an operand or an operator, and each (directed) edge $e \in \mathcal{E}$ indicates the control flow in the graph. Further, each node in the graph also specifies computation rules for the gradient of the outputs of the node with respect to the inputs to the node. Computational graphs can be nested and composed in about any manner, whilst preserving differentiability. An example computation graph for the function $3(xy + z)$ is shown in Fig.~\ref{fig:computational_graph_mockup}.

In a standard SLAM pipeline there are several subsystems/components that are not differentiable (i.e., for a few forward computations in the graph, gradients are unspecifiable). For example, in the context of dense 3D SLAM \cite{kinectfusion}\cite{pointfusion}, nonlinear least squares modules, raycasting routines, and discretizations are non-diffrentiable. Further, for several operations such as index selection / sampling, gradients exist, but are zero \emph{almost everywhere}, which result in extremely sparse gradient flows.

\subsection{\textbf{Method Overview}}

The objective of \gradslam{} is to make every computation in SLAM exactly realised as a composition of differentiable functions\footnote{Wherever exact differentiable realizations are not possible, we desire  \emph{as-exact-as-possible} realizations.}. Broadly, the sequence of operations in dense SLAM systems can be termed as \emph{odometry estimation} (frame-to-frame alignment), \emph{map building} (model-to-frame alignment/local optimization), and \emph{global optimization}. An overview of the approach is shown in \ref{fig:teaser_new}. 

% We assume that the SLAM system is operating on RGB-D images. The exposition also holds true for monocular images, as there are readily available approaches to predict per-pixel depth maps \cite{sfmlearner, neuralrgbd}\footnote{Understandably, such single-image depth prediction approaches do not generalize beyond the dataset they were trained on.}.

First, we provide a description of the precise issues that render nearly all of the aforementioned modules non-differentiable, and propose differntiable counterparts for each module. Finally, we show that the proposed differentiable variants allow the realization of several classic dense mapping algorithms (\emph{KinectFusion} \cite{kinectfusion}, \emph{PointFusion} \cite{pointfusion}, ICP-SLAM) in the \gradslam{} framework.\footnote{That is, realizable as \emph{fully} differentiable computational graphs.}
% We specifically show in Sec.~\ref{sec:case_studies} that three popular dense SLAM techniques (\emph{KinectFusion} \cite{kinectfusion}, \emph{PointFusion} \cite{pointfusion}, and ICP-SLAM \cite{ICP-SLAM}) can be realised as differentiable computational graphs. 

%%%%%%%%%%%%%%%%%%%%%%%%%%%%%% Grad-LM subsection
\subsection{\textbf{\gradlm{}: A differentiable nonlinear least squares solver}}
\label{sec:gradslam:gradlm}

\begin{figure}[!t]
    \centering
    \includegraphics[width=0.45\textwidth]{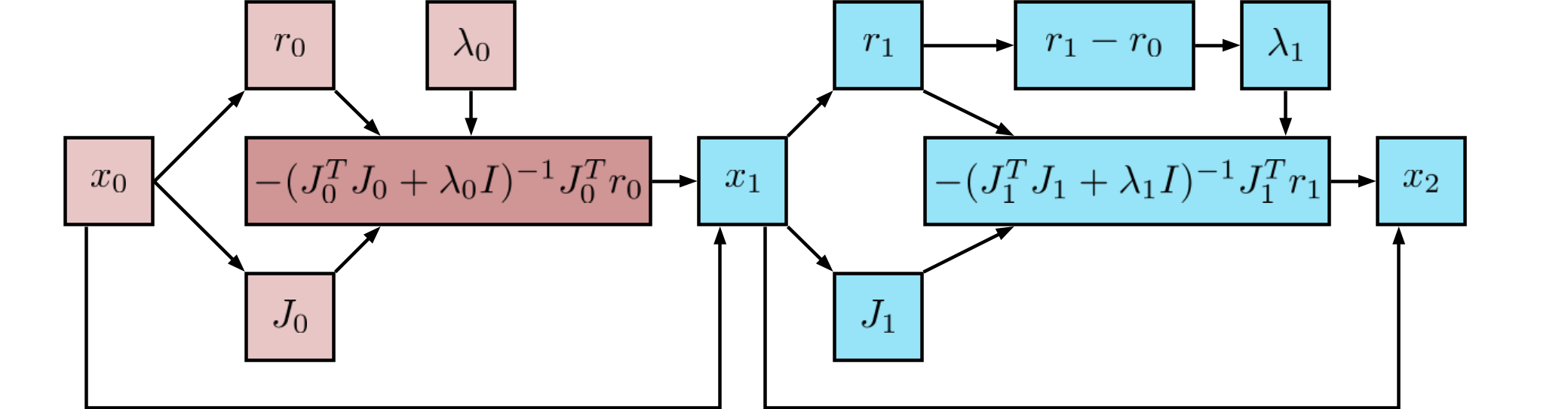}
    \caption{Computational graph for \emph{\gradlm{}}}
    \label{fig:gradLM}
\end{figure}

% \GI{I feel this paragraph can be reduced quite a bit, by explaining LM a little less}
% \iffalse
% Most state-of-the-art SLAM solutions optimize nonlinear least squares objectives to obtain local/globally consistent estimates of the robot state and the map. Such objectives are of the form $\mathbf{\frac{1}{2} \sum r(x)^2}$, where $\mathbf{r(x)}$ is a nonlinear function of residuals. 
% Example application scenarios that induce this nonlinear least squares form include visual odometry, depth measurement registration (e.g., ICP), and pose-graph optimization among others. Such objective functions are minimized using a succession of linear approximations ($\mathbf{r(x + \delta x)|_{x=x_0} = r(x_0) + J(x_0)\delta x}$), using Gauss-Newton (GN) or Levenberg-Marquardt (LM) solvers. While GN solvers are differentiable by unrolling the computational graph of the optimizer, they are i) numerically unstable when the Jacobian is near-singular, and ii) exhibit no convergence guarantees (in fact, they even diverge for a well-behaved class of functions \cite{RISE}). LM solvers trade-off convergence time for numerical stability and better guarantees. However, such \emph{trust-region} methods are not differentiable as they involve recalibration of optimizer parameters, based on a \emph{lookahead} operation over subsequent iterates \cite{dampingLM}.
% \fi

Most state-of-the-art SLAM solutions optimize nonlinear least squares objectives to obtain local/globally consistent estimates of the robot state and the map. Such objectives are of the form $\mathbf{\frac{1}{2} \sum r(x)^2}$, where $\mathbf{r(x)}$ is a nonlinear function of residuals. 
Example application scenarios that induce this nonlinear least squares form include visual odometry, depth measurement registration (e.g., ICP), and pose-graph optimization among others. Such objective functions are minimized using a succession of linear approximations ($\mathbf{r(x + \delta x)|_{x=x_0} = r(x_0) + J(x_0)\delta x}$), using Gauss-Newton (GN) or Levenberg-Marquardt (LM) solvers. GN solvers are extremely sensitive to intialization, numerical precision, and moreover, provide no guarantees on \emph{non-divergent} behavior. Hence most SLAM systems use LM solvers. 

\emph{Trust-region} methods (such as LM) are not differentiable as at each optimization step, they involve recalibration of optimizer parameters, based on a \emph{lookahead} operation over subsequent iterates \cite{dampingLM}. 
Specifically, after a new iterate is computed, LM solvers need to make a \emph{discrete} decision between damping or undamping the linear system. Furthermore, when undamping, the iterate must be restored to its previous value. This discrete \emph{switching} behavior of LM does not allow for gradient flow in the backward pass. 
% LS-Net \cite{lsnet} and BA-Net \cite{banet} have proposed learnable differentiable mappings within the LM routine. While LS-Net \cite{lsnet} altogether removes the LM-routine and drops in an LSTM (Long-Short Term Memory \cite{LSTM}) network to predict updates to the iterate, BA-Net \cite{banet} uses an LSTM network only to predict the damping coefficient. \LP{repeated from II-C} However, both these approaches require the LSTM networks to be pretrained, and this does not generalize to unseen data distributions.

\begin{figure}[!tbh]
    \centering
    \includegraphics[width=0.3\textwidth]{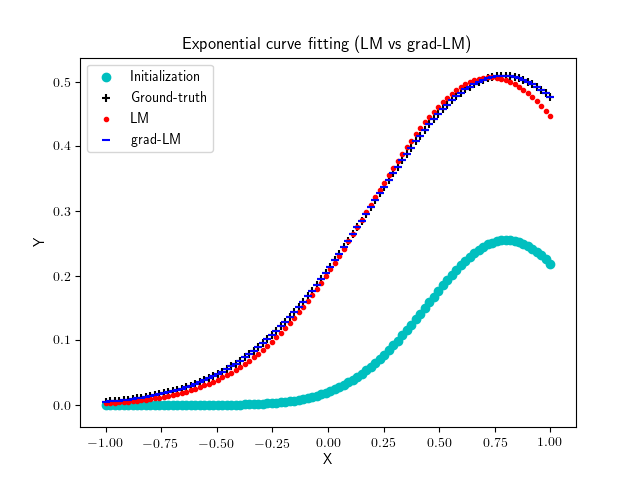}
    \caption{An example curve fitting problem, showing that \gradlm{} performs near-identical to LM, with the added advantage of being fully differentiable.}
    \label{fig:gradlm_curvefit}
\end{figure}

We propose a computationally efficient \emph{soft reparametrization} of the damping mechanism to enable differentiability in LM solvers. Our key insight is that, if $\mathbf{r_0 = r(x_0)^Tr(x_0)}$ is the norm of the error at the current iterate, and $\mathbf{r_1 = r(x_1)^Tr(x_1)}$ is the norm of the error at the \emph{lookahead} iterate, the value of $\mathbf{r_1 - r_0}$ determines whether to damp or to undamp. And, only when we choose to undamp, we revert to the current iterate. We define two smooth \emph{gating} functions $Q_x$ and $Q_\lambda$ based on the generalized logistic function \cite{richards1959flexible} to update the iterate and determine the next damping coefficient.
\begin{equation}
\begin{split}
    \lambda_1 = Q_\lambda(r_0, r_1) & = \lambda_{min} + \frac{\lambda_{max} - \lambda_{min}}{1 + De^{-\sigma (r_1 - r_0)}} \\
    Q_x(r_0, r_1) & = x_0 + \frac{\delta x_0}{1 + e^{-(r_1 - r_0)}}
\end{split}
\label{eqn:gradlm-gating}
\end{equation}
\noindent
where $D$ and $\sigma$ are tunable parameters that control the falloff \cite{richards1959flexible}. Also $[\lambda_{min}, \lambda_{max}]$ is the range of values the damping function can assume. Notice that this smooth parameterization of the LM update allows the optimizer to be expressed as a fully differentiable computational graph (Fig.~\ref{fig:gradLM}).

It must be noted that this scheme can be modified to accommodate other kinds of gating functions, such as hyperbolic curves. We however, choose the above gating functions, as they provide sufficient flexibility. A thorough treatment of the impact of the choice of gating functions on performance is left for future work.

\subsection{\textbf{Differentiable mapping}}
\label{sec:gradslam:mapping}

\begin{figure}[!t]
    \centering
    \includegraphics[width=\columnwidth]{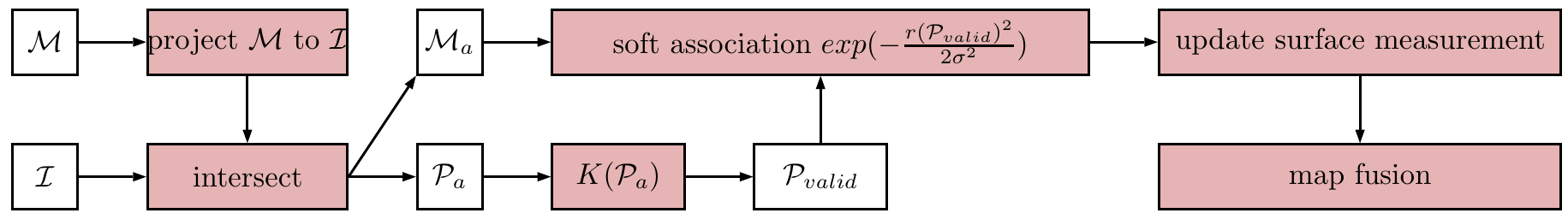}
    \caption{Computation graph for the differentiable mapping module. The uncolored boxes indicate intermediate variables, while the colored boxes indicate processing blocks. Note that the specific choice of the functions for \texttt{update surface measurement} and \texttt{map fusion} depend on the map representation used.}
    \label{fig:gradMap}
\end{figure}

Another non-smooth operation in dense SLAM is map construction (\emph{surface measurement}). For example, consider a \emph{global} map $\mathcal{M}$ being built in the reference frame of the first image-sensor measurement $I_0$. 
When a new frame $I_k$ arrives at time $k$, dense SLAM methods need to align the surface measurement being made in the live frame, with the map $\mathcal{M}$. Notwithstanding the specific choice of map representation (i.e., pointclouds, signed-distances, surfels), a generic \emph{surface alignment} process comprises the following steps. 
\begin{enumerate}
    \item The map $\mathcal{M}$ is intersection-tested with the live frame, to determine the \emph{active set} $\mathcal{M}_a$ of map elements, and the active set of image pixels $\mathcal{P}_a$. The remaing map elements are \emph{clipped}.
    \item Active image pixels $\mathcal{P}_a$ are checked for measurement validity (e.g., missing depth values / blurry pixels, etc.). This results in a \emph{valid active set} of image pixls $\mathcal{P}_{valid}$
    \item The set of pixels in $\mathcal{P}_{valid}$ is backprojected to 3D and compared with the map. At this stage, it must be discerned whether these pixels measure existing elements in $\mathcal{M}_a$, or if they measure a new set of elements that need to be added to the global map. 
    \item Once the above decision is made, these surface measurements are \emph{fused} into the global map. The choice of the fusion mechanism is dependent on the underlying representation of each map element (points, surfels, TSDF, etc.).
\end{enumerate}

The above process involves a number of differentiable yet non-smooth operations (clipping, indexing, thresholding, new/old decision, active/inactive decision, etc.). Although the above sequence of operations can be represented as a computation graph, it will not necessarily serve our purpose here since, even though (local) derivatives can be defined for operations such as clipping, indexing, thresholding, and discrete decisions, these derivatives exist only at that single point. The overall function represented by the computation graph will  have undefined gradients "almost everywhere" (akin to step functions). We propose to mitigate this issue by making the functions locally \emph{smooth}. Concretely, we propose the following corrective measures.

\begin{enumerate}
    \item The surface measurement made by each valid pixel $p$ in the live frame (i.e., $p \in \mathcal{P}_{valid}$) is not a function of $p$ alone. Rather, it is the function of the pixel $p$ and its (active/inactive) neighbours $nbd(p)$, as determined by a \emph{kernel} $K(p,nbd(p))$.
    \item When a surface measurement is transformed to the global frame, rather than using a \emph{hard} (one-one) association between a surface measurement and a map element, we use a \emph{soft} association to multiple map elements, in accordance with the sensor characteristics.
    \item Every surface measurement is, by default, assumed to represent a new map element, which is passed to a \emph{differentiable fusion} step (\cf{} Sec~\ref{sec:gradSLAM:fusion}).
\end{enumerate}

The kernel $K(p, nbd(p))$ can be a discrete approximation (e.g., constant within a pixel) or can vary at the subpixel level, based on the choice of the falloff function. For faster computation and coarse gradients, we use a bilinear interpolation kernel. While bilinear interpolation is a sensible approximation for image pixels, this is often a poor choice for use in 3D \emph{soft} associations. For forming 3D associations, we leverage characteristics of RGB-D sensors in defining the soft falloff functions. Specifically, we compute, for each point $P$ in the live surface measurement, a set of closest candidate points in a region $exp\left(-\frac{r(P)^2}{2\sigma^2}\right)$, where $r(P)$ is the radial depth of the point from the camera ray, and $\sigma$ affects the falloff region.\footnote{This is a well-known falloff function, usually with Kinect-style depth sensors \cite{pointfusion, kinectnoise, curless1996volumetric}.}
%%%%%%%%%%%%%%%%%%%%%%%%%%%%%%

%%%%%%%%%%%%%%%%%%%%%%%%%%%%%% Fusion subsection
\subsection{Differentiable map fusion}
\label{sec:gradSLAM:fusion}

The aforementioned differentiable mapping strategy, while providing us with a smooth observation model, also causes an undesirable effect: the number of map elements increases in proportion with exploration time. However, map elements should ideally increase with proportion to the \emph{explored volume of occupied space}, rather than with exploration time. Conventional dense mapping techniques (e.g., KinectFusion \cite{kinectfusion}, PointFusion \cite{pointfusion}) employ this through \emph{fusion} of redundant observations of the same map element. As a consequence, the recovered map has a more manageable size, but more importantly, the reconstruction quality improves greatly. While most fusion strategies are differentiable (eg. \cite{kinectfusion, pointfusion}), they impose falloff thresholds that cause an abrupt change in gradient flow at the truncation point. We use a logistic falloff function, similar to Eq.~\ref{eqn:gradlm-gating}, to ease gradient flow through these truncation points.
% \TODO{Plot a 1D truncation function and its derivative. Plot a 1D truncation function with our modified falloff and its derivative. (This is probably better suited for the KF case study?)}
%%%%%%%%%%%%%%%%%%%%%%%%%%%%%%

%%%%%%%%%%%%%%%%%%%%%%%%%%%%%% Ray subsection
\subsection{Differentiable ray backprojection}
\label{sec:gradSLAM:ray}

%\TODO{Move this into KinectFusion case-study?? As this is only applicable for the voxel reconstruction setup.}

\begin{figure}[!htb]
    \centering
    \includegraphics[width=\columnwidth]{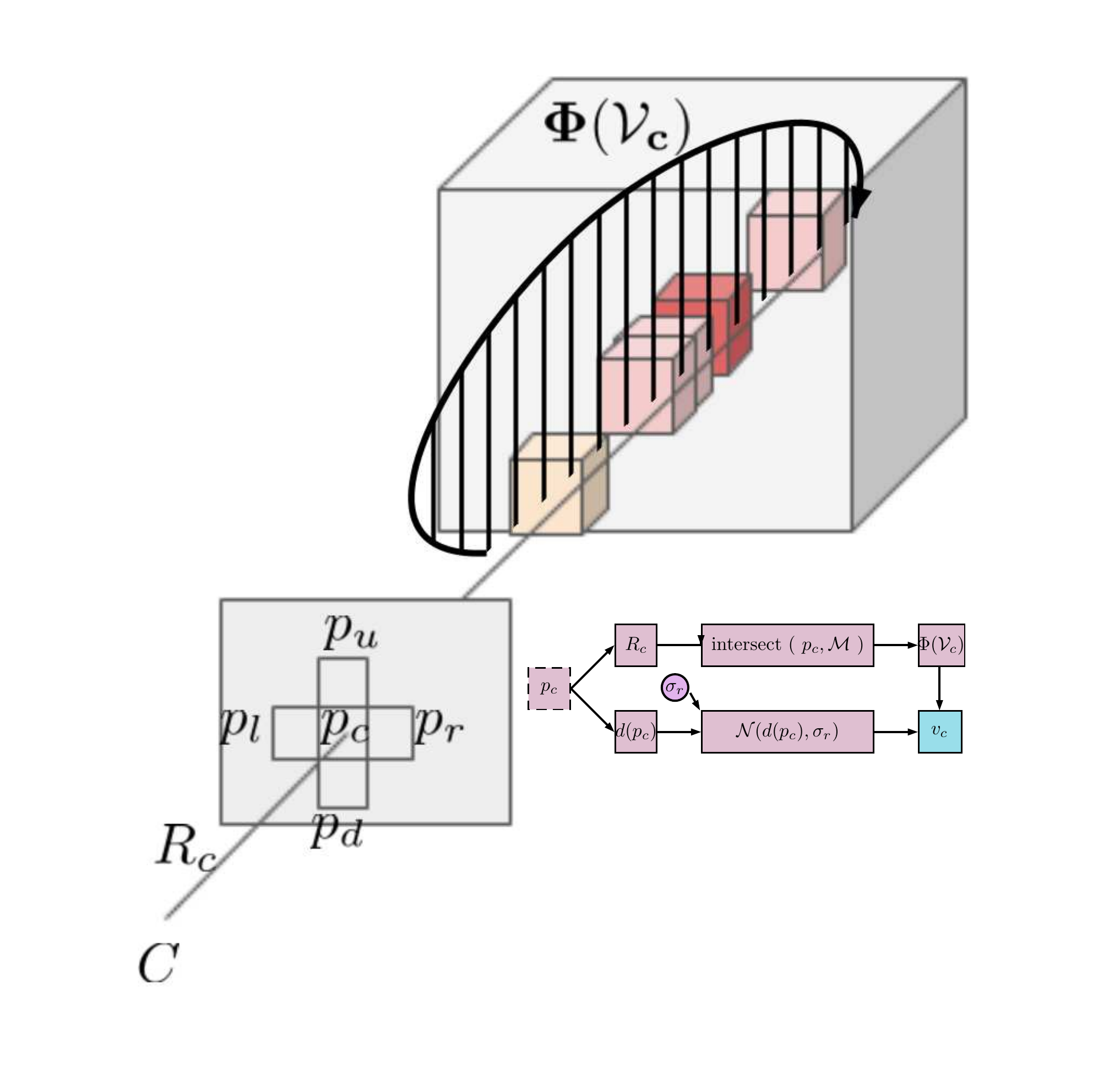}
    % \vspace{-0.6cm}
    \caption{\textbf{Ray differentials}: Inset shows the computation graph of the ray value computation. The dashed rectangle is not differentiable, and its derivatives are approximated as shown in Eq~\ref{eq:ray_derivative}}
    \label{fig:ray}
    % \vspace{-0.3cm}
\end{figure}

Some dense SLAM systems \cite{kinectfusion,kintinuous} perform global pose estimation by raycasting a map to a live frame. Such an operation inherently involves non-differentiable steps. First, from each pixel in the image, a ray from the camera is backprojected, and its intersection with the first map element along the direction of the ray is determined. This involves marching along the ray until a map element is found, or until we exit the bounds of reconstruction. 
Usual (non-differentiable) versions of ray marching use max-min acceleration schemes \cite{max-min-marching} or rely on the existence of volumetric signed distance functions \cite{kinectfusion}. Several attempts have been made to make the raycasting operation differentiable. Scene representation networks \cite{scene_rep_nets} proposes to predict ray marching steps using an LSTM. 
In other works such as DRC \cite{drc} and WS-GAN \cite{wsgan}, the authors pool over all voxels along a ray to compute the \emph{potential} of a ray. 
In this work, we make one enhancement to the ray pooling operation. We pool over all voxels along a ray, but have a Gaussian falloff defined around the depth measurement of the image pixel through which the ray passes. Further, we use finite differences to compute the derivative of the ray potential with respect to the pixel neighbourhood. We use the finite differences based ray differentials defined in Igehy \etal{} \cite{ray_differentials}. 
If $p_{c}$ is the image pixel that the ray $R_c$ \emph{pierces}, and $\mathcal{V}_c = \{v_c\}$ is the set of all voxels it pierces, then the \emph{aggregated value} of the ray is denoted $v_c$ (with respect to an \emph{aggregation} function $\Phi(\psi(v_c) \  \forall v_c \in \mathcal{V}_c)$). 
The aggregation function simply multiplies each value $\psi(v_c)$ with the density of the Gaussian fallof at $v_c$, and normalizes them. Similarly $v_l$, $v_r$, $v_u$, and $v_b$ are the \emph{aggregated values} of rays emanating from the pixels to the left, right, above, and below $p_c$ respectively. Then, the partial derivative $\frac{\partial v_c}{\partial c}$ can be approximated as

\begin{equation}
    \frac{\partial v_c}{\partial p_c} = \begin{pmatrix} (v_r - v_l) / 2 \\ (v_u - v_d) / 2 \end{pmatrix}
    \label{eq:ray_derivative}
\end{equation}

An illustration of the ray differential computation scheme can be found in Fig.~\ref{fig:ray}.

\section{Case Studies: KinectFusion, PointFusion, and \emph{ICP-SLAM}}
\label{sec:case_studies}

To demonstrate the applicability of the \gradslam{} framework, we leverage the differentiable computation graphs specified in Sec~\ref{sec:gradslam} and compose them to realise three practical SLAM solutions. In particular, we implement differentiable versions of the \emph{KinectFusion} \cite{kinectfusion} algorithm that constructs TSDF-based volumetric maps, the \emph{PointFusion} \cite{pointfusion} algorithm that constructs surfel maps, and a pointcloud-only SLAM framework that we call \emph{ICP-SLAM}.

\subsection{KinectFusion}
\label{sec:case_studies:kf}

Recall that KinectFusion \cite{kinectfusion} alternates between \emph{tracking} and \emph{mapping} phases. In the tracking phase, the entire up-to-date TSDF volume is raycast onto the live frame, to enable a point-to-plane ICP that aligns the live frame to the raycast model. Subsequently, in the mapping phase, surface measurements from the current frame are \emph{fused} into the volume, using the TSDF fusion method proposed in \cite{kinectfusion}. The surface measurement is given as (\cf \cite{kinectfusion})
% \vspace{-0.4cm}

\begin{equation}
\begin{split}
    sdf(p) & = trunc( \| K^{-1} x \|_2^{-1} \| t - p \|_2 - depth(x) ) \\ 
    trunc(sdf) & = min(1, \frac{sdf}{\mu})(sign(sdf)) \ \ \ iff \ sdf \  \ge -\mu
\end{split}
% \vspace{-0.25cm}
\end{equation}

Here, $p$ is the location of a voxel in the camera frame, and $x = \lfloor \pi (K p) \rfloor$ is the live frame pixel to which $p$ projects to. $\mu$ is a parameter that determines the threshold beyond which a surface measurement is invalid. However, we note that the \emph{floor} operator is non-differentiable "almost everywhere". Also, the truncation operator, while differentiable within a  distance of $\mu$ from the surface, is abruptly truncated, which hinders gradient flow . Instead, we again use a generalized logistic function \cite{richards1959flexible} to create a smooth truncation, which provides better-behaved gradients at the truncation boundary. The other steps involved here, such as raycasting, ICP, etc. are already differentiable in the \gradslam{} framework (\cf Sec~\ref{sec:gradslam}). Fusion of surface measurements is perfomed using the same approach as in \cite{kinectfusion} (weighted averaging of TSDFs).
%(\TODO{Make a 1D TSDF example, plot its gradient, and plot the gradient with our chosen falloff)}

% \vspace{-0.25cm}
\subsection{PointFusion}
\label{sec:case_studies:pf}

As a second example, we implement PointFusion \cite{pointfusion}, which incrementally fuses surface measurements to obtain a global surfel map. Surfel maps compare favourably to volumetric maps due to their reduced memory usage.\footnote{On the flipside, surfel-based algorithms are harder to parallelize compared to volumetric fusion.} We closely follow our differentiable mapping formulation (\cf Sec~\ref{sec:gradslam:mapping}) and use surfels as map elements. We adopt the fusion rules from \cite{pointfusion} to perform map fusion.

% \vspace{-0.25cm}
\subsection{\emph{ICP-SLAM}}

As a baseline example, we implement a simple pointcloud based SLAM technique, which uses ICP to incrementally register pointclouds to a global pointcloud set. In particular, we implement two systems. The first one aligns every pair of consecutive incoming frames, to obtain an odometry estimate (also referred to as \emph{frame-to-frame alignment} or ICP-Odometry). The second variant performs what we call \emph{frame-to-model alignment} (ICP-SLAM). That is, each incoming frame is aligned (using ICP) with a pointcloud containing the entire set of points observed thus far.

\begin{table*}[!ht]
\begin{adjustbox}{max width=\linewidth}
\begin{tabular}{|c|c|c|c|c||c|c|c|c||c|c|c|c|}
\hline
\multicolumn{1}{|c|}{$T_{max}$ = $10$ iters}     & \multicolumn{4}{c|}{\textit{Exponential}}                                                                   & \multicolumn{4}{c|}{\textit{Sine}}                                                                           & \multicolumn{4}{c|}{\textit{Sinc}}                                                                          \\ \hline
\multicolumn{1}{|c|}{}                     & \multicolumn{1}{c|}{\textbf{GD}} & \multicolumn{1}{c|}{\textbf{GN}} & \multicolumn{1}{c|}{\textbf{LM}} & \multicolumn{1}{c||}{\textbf{\gradlm{}}} & \multicolumn{1}{c}{\textbf{GD}} & \multicolumn{1}{c|}{\textbf{GN}} & \multicolumn{1}{c|}{\textbf{LM}} & \multicolumn{1}{c||}{\textbf{\gradlm{}}} & \multicolumn{1}{c|}{\textbf{GD}} & \multicolumn{1}{c|}{\textbf{GN}} & \multicolumn{1}{c|}{\textbf{LM}} & \multicolumn{1}{c|}{\textbf{\gradlm{}}} \\ \hline
$\|a_{pred} - a_{gt}\|^2$                                        & $\mb{0.422}$                  & $0.483$                  & $0.483$                  & $0.483$                   & $0.379$                  & $\mb{0.341}$                  & $0.342$                  & $0.342$                   & $2.929$                  & $0.304$                  & $0.304$                   & $\mb{0.304}$                   \\
$\|t_{pred} - t_{gt}\|^2$                                        & $0.606$                  & $\mb{0.50}$                   & $0.550$                  & $0.550$                   & $\mb{0.222}$                  & $0.359$                  & $0.360$                  & $0.360$                   & $3.024$                  & $0.304$                  & $0.304$                  & $\mb{0.040}$                    \\
$\|w_{pred} - w_{gt}\|^2$                                        & $1.268$                  & $0.667$                  & $0.075$                  & $\mb{0.075}$                   & $1.215$                  & $\mb{0.080}$                  & $0.084$                  & $0.085$                   & $0.462$                  & $\mb{10^{-7}}$                   & $0.023$                  & $10^{-4}$                    \\ \hline
$\|f(x)_{pred} - f(x)_{gt}\|^2$                     & $0.716$                  & $0.160$                  & $0.163$                  & $\mb{0.160}$                   & $0.666$                  & $0.148$                  & $0.152$                  & $\mb{0.148}$                   & $0.700$                  & $\mb{5 \times 10^{-8}}$                   & $0.005$                  & $4 \times 10^{-5}$                    \\ \hline \hline
$T_{max}$ = $50$ iters                          &                        &                        &                        &                         &                        &                        &                        &                         &                        &                        &                        &                         \\ \hline
$\|a_{pred} - a_{gt}\|^2$                                        & $0.365$                  & $0.275$                  & $\mb{0.231}$                  & $0.275$                   & $0.486$                  & $\mb{0.429}$                  & $0.434$                  & $0.434$                   & $3.329$                  & $0.380$                  & $0.380$                  & $\mb{0.380}$                   \\
$\|t_{pred} - t_{gt}\|^2$                                        & $0.263$                  & $0.219$                  & $0.231$                  & $\mb{0.218}$                   & $0.519$                  & $\mb{0.455}$                  & $0.459$                  & $0.460$                   & $2.739$                  & $0.380$                  & $0.380$                  & $\mb{0.380}$                   \\
$\|w_{pred} - w_{gt}\|^2$                                        & $1.220$                  & $0.205$                  & $\mb{0.007}$                  & $0.369$                   & $1.327$                  & $0.273$                  & $\mb{0.376}$                  & $0.383$                   & $0.383$                  & $\mb{2 \times 10^{-7}}$                   & $0.202$                 & $4 \times 10^{-5}$                    \\ \hline
$\|f(x)_{pred} - f(x)_{gt}\|^2$                     & $0.669$                  & $0.083$                  & $\mb{0.004}$                  & $0.078$                   & $0.673$                  & $0.153$                  & $0.153$                  & $\mb{0.151}$                   & $0.795$                  & $\mb{2 \times 10^{-7}}$                   & $0.005$                  & $3 \times 10^{-5}$                    \\ \hline \hline
$T_{max}$ = $100$ iters                         &                        &                        &                        &                         &                        &                        &                        &                         &                        &                        &                        &                         \\ \hline
$\|a_{pred} - a_{gt}\|^2$                                        & $\mb{0.431}$                  & $0.475$                  & $0.480$                  & $0.487$                   & $0.486$                  & $\mb{0.429}$                  & $0.434$                  & $0.434$                   & $2.903$                  & $0.196$                  & $0.196$                  & $\mb{0.196}$                   \\
$\|t_{pred} - t_{gt}\|^2$                                        & $0.466$                  & $\mb{0.311}$                  & $0.378$                  & $0.323$                   & $0.519$                  & $\mb{0.455}$                  & $0.459$                  & $0.460$                   & $2.847$                  & $0.196$                  & $0.196$                  & $\mb{0.196}$                   \\
$\|w_{pred} - w_{gt}\|^2$                                       & $1.140$                  & $0.364$                  & $0.066$                  & $\mb{0.065}$                   & $1.327$                  & $\mb{0.273}$                  & $0.376$                  & $0.382$                   & $0.601$                  & $\mb{10^{-7}}$                   & $0.026$                  & $9 \times 10^{-5}$                    \\ \hline
$\|f(x)_{pred} - f(x)_{gt}\|^2$ & $0.662$                  & $0.243$                  & $\mb{0.162}$                  & $0.230$                   & $0.673$                  & $0.153$                  & $0.153$                  & $\mb{0.151}$                   & $0.707$                  & $\mb{6 \times 10^{-8}}$                   & $0.005$                  & $4 \times 10^{-5}$   \\ \hline
\end{tabular}
\end{adjustbox}
\label{table:gradlm-toy-suite}
\caption{\textbf{\gradlm{}} performs quite similarly to its non-differentiable counterpart, on a variety of non-linear functions, and at various stages of optimization. Here, \textbf{GD}, \textbf{GN}, and \textbf{LM} refer to gradient descent, Gauss-Newton, and Levenberg-Marquardt optimizers respectively.}
\end{table*}

% \vspace{-0.25cm}
\section{Experiments and results}
\label{sec:results}

\subsection{Differentiable optimization}

% \TODO{Refine this: just putting in number for now.}

In Sec~\ref{sec:gradslam:gradlm}, we introduced two generalized logistic functions $Q_\lambda$ and $Q_x$ to compute the damping functions as well as the subsequent iterates. We conduct multiple experiments to verify the impact of this approximation on the performance (convergence speed, quality of solution) of nonlinear least squares solvers. 

We first design a test suite of nonlinear curve fitting problems (inspiration from \cite{lsnet}), to measure the performance of \gradlm{} to its non-differentiable counterpart. We consider three nonlinear functions, \emph{viz.} \emph{exponential}, \emph{sine}, and \emph{sinc}, each with three parameters $a, t,$ and $w$. 

\begin{equation}
    \begin{aligned}
        f(x) & = a \  exp\left(-\frac{(x - t)^2}{2w^2}\right) \\
        f(x) & = sin(ax + tx + w) \\
        f(x) & = sinc(ax + tx + w)
    \end{aligned}
\end{equation}

For each of these functions, we uniformly sample the parameters $p = \{a, t, w\}$ to create a suite of ground-truth curves, and uniformly sample an initial guess $p_0 = \{a_0, t_0, w_0\}$ in the interval $\left[-6, 6\right]$. We sample $100$ problem instances for each of the three functions. We run a variety of optimizers (such as gradient descent (GD), Gauss-Newton (GN), LM, and \gradlm{}) for a maximum of $10, 50,$ and $100$ iterations. We compute the mean squared error in \emph{parameter space} (independently for each parameter $a, t, w$) as well as in \emph{function space} (i.e., $\|f(x)_{pred} - f(x)_{gt}\|^2$. Note that these two errors are not necessarily linearly related, as the interaction between the parameters and the function variables are highly nonlinear. The results are presented in Table~\ref{table:gradlm-toy-suite}. It can be seen that \gradlm{} performs near-identically to LM.

\begin{figure*}[!h]
    \centering
    \includegraphics[width=\textwidth]{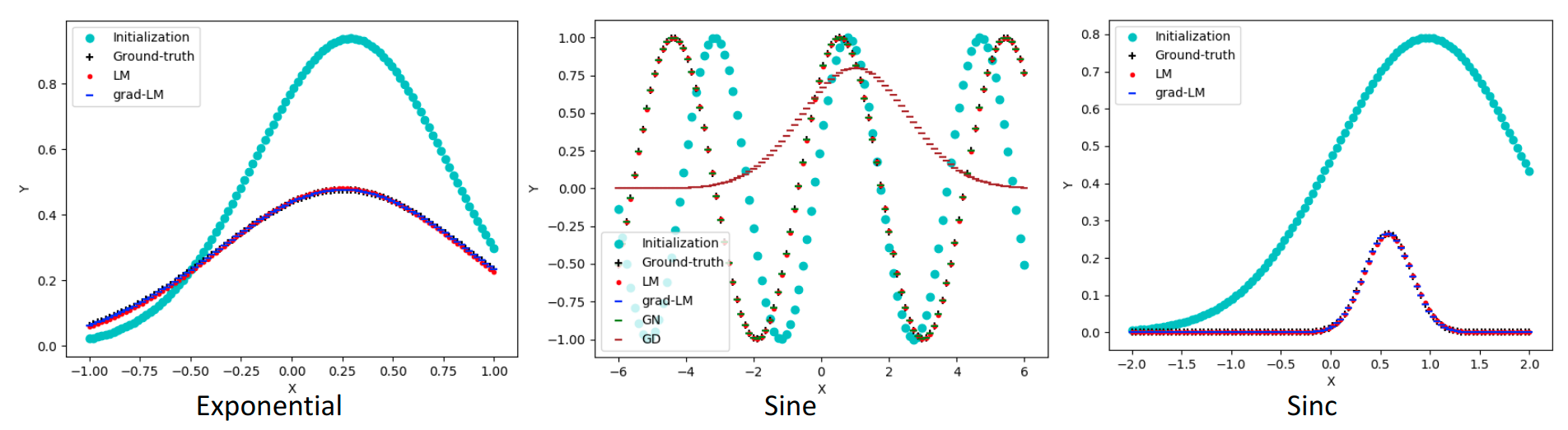}
    \caption{\textbf{\gradlm{}} performs comparably to LM optimizers. In this figure, we show example curve fitting problems from the test suite.}
    \label{fig:my_label}
\end{figure*}

\subsection{Comparitive analysis of case studies}
% \vspace{-0.15cm}

In Sec~\ref{sec:case_studies}, we implemented KinectFusion \cite{kinectfusion}, PointFusion \cite{pointfusion}, and ICP-SLAM as differentiable computational graphs. Here, we present an analysis of how each of the approaches compare to their non-differentiable counterparts. Table~\ref{tab:gradslam_performance} shows the trajectory tracking performance of the non-differentiable and differentiable ($\nabla$) versions of ICP-Odometry, ICP-SLAM, and PointFusion. We observe no virtual change in performance when utilizing the differentiable mapping modules and \gradlm{} for optimization. This is computed over split subsets of the \texttt{living\_room\_traj0} sequence.

\begin{figure*}[!htb]
    \centering
    \includegraphics[width=\textwidth]{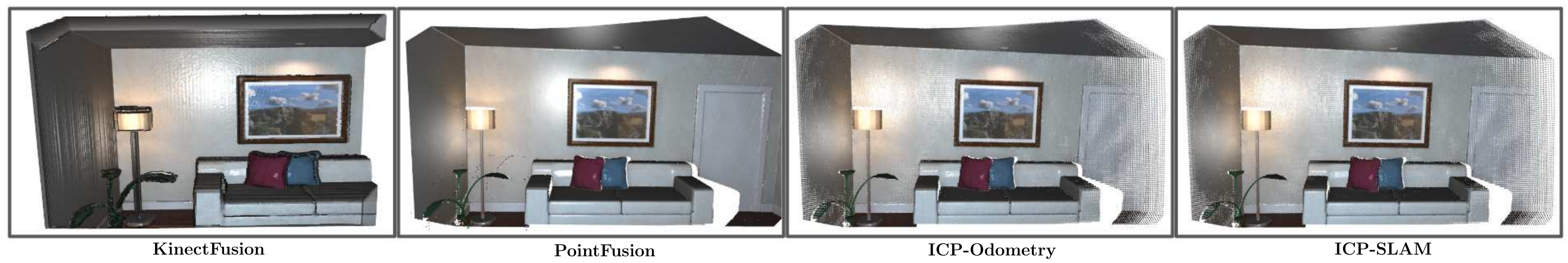}
    \caption{\textbf{Qualitative results}: On the \textbf{\texttt{living room lr kt0}} sequence of the ICL-NUIM dataset \cite{icl-nuim}. The reconstructions are near-identical to their non-differentiable counterparts. However, distinct from classic SLAM approaches, these reconstructions allow for gradients to flow from a 3D map element all the way to the entire set of pixel-space measurements of that element.}
    \label{fig:qualitative}
    % \vspace{-0.7cm}
\end{figure*}

We also evaluate the reconstruction quality of $\nabla$-KinectFusion with that of Kintinuous \cite{kintinuous}. On a subsection of the \texttt{living\_room\_traj0} sequence of the ICL-NUIM \cite{icl-nuim} benchmark, the surface reconstruction quality of Kintinuous is $18.625$, while that of differentiable KinectFusion is $21.301$ (better). However, this quantity is misleading, as Kintinuous only retains a subset of high confidence points in the extracted mesh, while our differentiable KinectFusion outputs (see Fig.~\ref{fig:qualitative}) contain a few noisy artifacts, due to our smooth truncation functions.

\begin{table}[!tbh]
    \centering
    \begin{adjustbox}{max width=\linewidth}
    \begin{tabular}{c|c|c}
        \hline
        \textbf{Method} & \textbf{ATE} & \textbf{RPE} \\
        \hline
        ICP-Odometry (non-differentiable) & $0.029$ & $0.0318$\\
        $\nabla$ICP-Odometry & $\mathbf{0.01664}$ & $\mathbf{0.0237}$\\
        \hline
        ICP-SLAM (non-differentiable) & $0.0282$ & $0.0294$\\
        $\nabla$ICP-SLAM & $\mathbf{0.01660}$ & $\mathbf{0.0204}$\\
        \hline
        PointFusion (non-differentiable) & $\mathbf{0.0071}$ & $\mathbf{0.0099}$\\
        $\nabla$PointFusion & $0.0072$ & $0.0101$\\
        \hline
        KinectFusion (non-differentiable) & $\mathbf{0.013}$ & $\mathbf{0.019}$ \\
        $\nabla$KinectFusion & $0.016$ & $0.021$ \\ \hline
    \end{tabular}
     \end{adjustbox}
    \caption{\textbf{Performance of \gradslam{}}. The differentiable counterparts perform nearly similar to their non-differentiable counterparts (ATE: Absolute Trajectory Error, RPE: Relative Pose Error).}
    \label{tab:gradslam_performance}
    % \vspace{-1cm}
\end{table}

\subsection{Qualitative results}

\gradslam{} works out of the box on multiple other RGB-D datasets. Specifically, we present qualitative results of running our differentiable SLAM systems on RGB-D sequences from the TUM RGB-D dataset~\cite{tumrgbd}, ScanNet~\cite{scannet}, as well as on an in-house sequence captured from an Intel RealSense D435 camera.

\begin{figure}[!htb]
    \centering
    \includegraphics[width=0.5\textwidth]{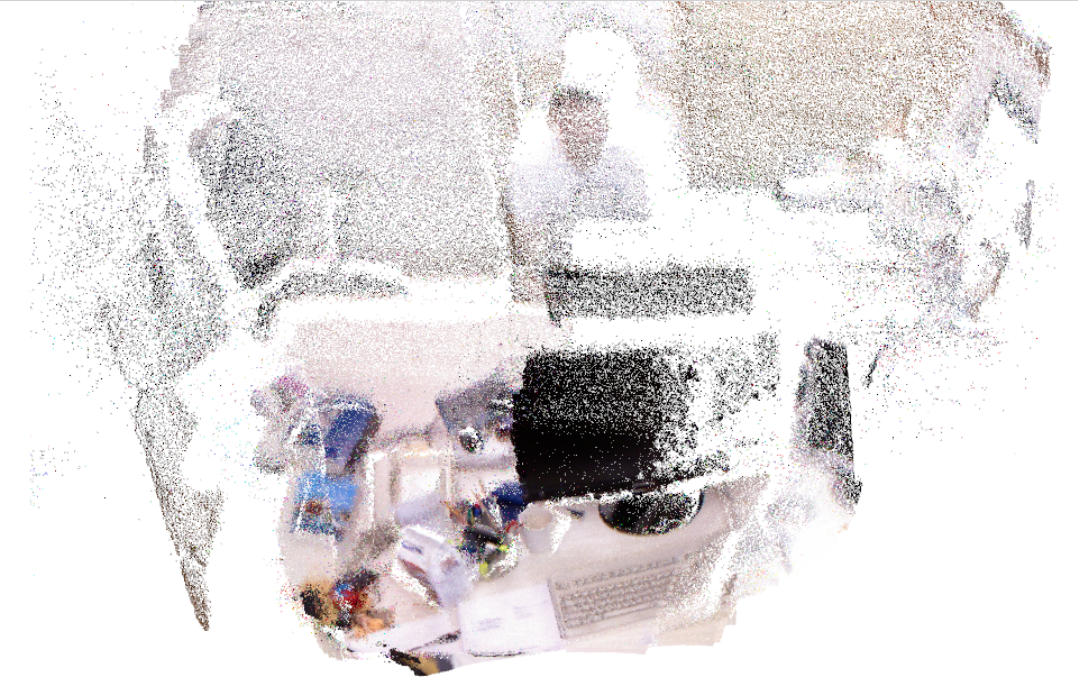}
    \caption{Reconstruction obtained upon running the differentiable ICP-Odometry pipeline on a subsection of the \texttt{rgbd\_dataset\_freiburg1\_xyz} sequence.}
    \label{fig:tum}
\end{figure}

\begin{figure*}[!htb]
    \centering
    \includegraphics[width=\textwidth]{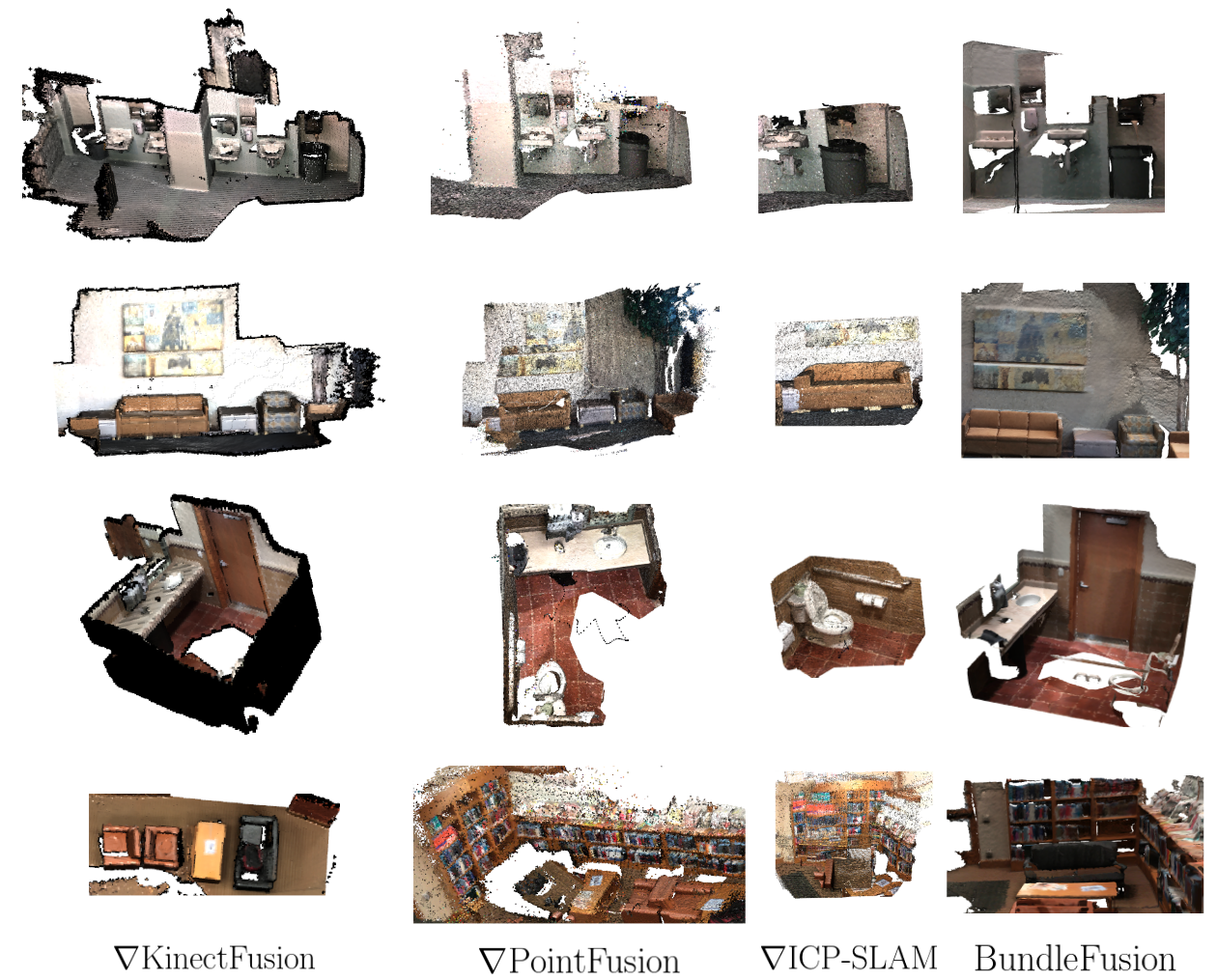}
    \caption{\textbf{Qualitative results} on sequences from the ScanNet~\cite{scannet} dataset. Owing to GPU memory constraints, we use each of the differentiable SLAM systems ($\nabla$KinectFusion, $\nabla$PointFusion, and $\nabla$ICP-SLAM) to reconstruct parts of the scene. We also show outputs from BundleFusion~\cite{bundlefusion} for reference.}
    \label{fig:scannet}
\end{figure*}

\begin{figure*}[!htb]
    \centering
    \includegraphics[width=0.8\textwidth]{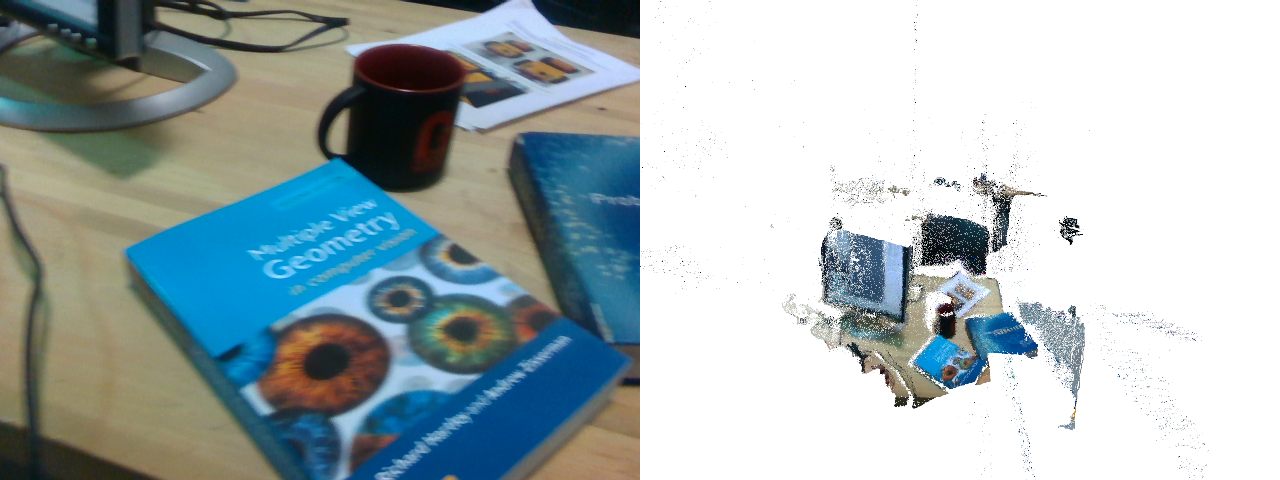}
    \caption{\textbf{In-house sequence} collected from an Intel RealSense D435 camera. The reconstruction (right) is obtained by running $\nabla$PointFusion. Note that we do not perform any noise removal. Differentiable noise filtering is left for future work.}
    \label{fig:mrsd}
\end{figure*}

Fig.~\ref{fig:tum}-~\ref{fig:mrsd} show qualitative results obtained by running \gradslam{} on a variety of sequences from the TUM RGB-D benchmark (Fig.~\ref{fig:tum}), ScanNet (Fig.~\cite{scannet}), and an in-house sequence (Fig.~\ref{fig:mrsd}). These differentiable SLAM systems all execute fully on the GPU, and are capable of computing gradients with respect to \emph{any} intermediate variable (Eg. camera poses, pixel intensities/depths, optimization parameters, camera intrinsics, etc.).

\subsection{Analysis of Gradients}
\label{sec:results:depthgrad}
\begin{figure*}[!t]
    \centering
    \includegraphics[width=\textwidth]{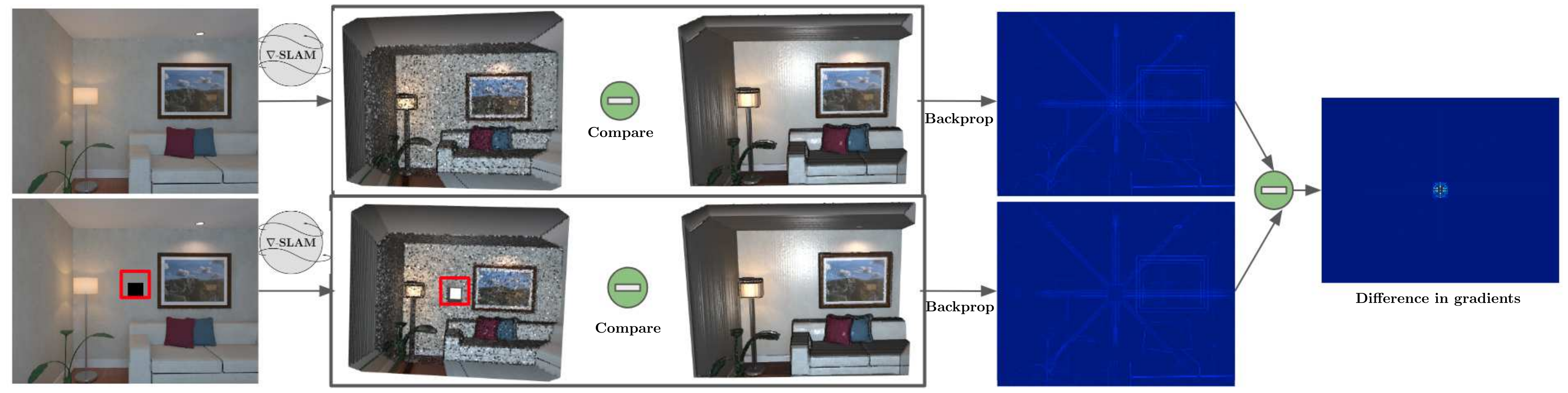}
    \caption{\textbf{Analysis of gradients}: \gradslam{} enables gradients to flow all the way back to the input images. \emph{Top}: An RGB-D image pair (depth not shown) is passed through \gradslam{}, and reconstruction error is computed using a precise fused map. Backpropagation passes these gradients all the way back to the depth map (blue map). \emph{Bottom}: An explicit occluder added to the center of the RGB-D pair. This occluder distorts the construction by creating a gaping hole through it. But, using the backpropagated gradients, one can identify the set of image/depthmap pixels that result in a particular area to be reconstructed imperfectly.}
    \label{fig:depthgrad}
    % \vspace{-0.75cm}
\end{figure*}

% \vspace{-0.15cm}
The computational graph approach of \gradslam{} allows us to recover meaningful gradients of 2D (or 2.5D) measurements with respect to a 3D surface reconstruction. 
In Fig.~\ref{fig:depthgrad}, the top row shows an RGB-D image differentiably transformed---using \gradslam{}---into a (noisy) TSDF surface measurement, and then compared to a more precise global TSDF map. Elementwise comparision of aligned volumes gives us a reconstruction error, whose gradients are backpropagated through to the input depthmap using the computational graph maintained by \gradslam{} (and visualized in the depth image space). 
%can then be used to analyze the gradients of each pixel in the vertex and depth maps with respect to this volumetric error. 
In the second row, we intentionally introduce an occluder that masks out a small ($40 \times 40$) region in the RGB-D image, thereby introducing a reconstruction artifact. Computing the volumetric error between the global and local occluded TSDF volumes and inspecting the gradients with respect to the input indicates the per pixel contribution of the occluding surface to the volumetric error. \emph{Thus, \gradslam{} provides a rich interpretation of the computed gradients: they denote the contribution of each pixel towards the eventual 3D reconstruction.}

\subsection{Application: RGB and depth completion}

\begin{figure*}[tb]
    \centering
    \includegraphics[width=0.95\textwidth]{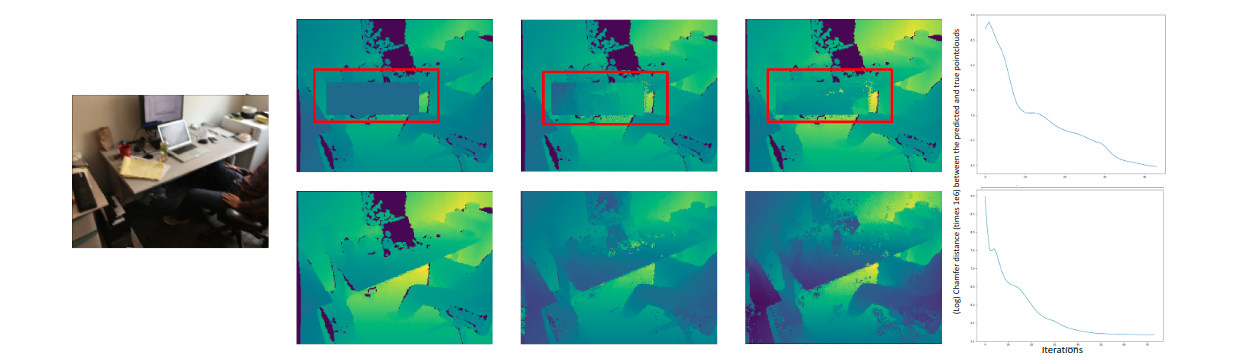}
    \caption{\textbf{End-to-end gradient propagation}: (\emph{Top}): A chunk of a depth map is \emph{chopped}. The resultant sequence is reconstructed using $\nabla$PointFusion and the pointcloud is compared to a \emph{clean} one reconstructed using the unmodified depth map. The Chamfer distance between these two pointclouds is used to define a reconstruction error between the two clouds, which is backpropagated through to the input depth map and  updated by gradient descent. (\emph{Bottom}): Similar to the Fig.~\ref{fig:depthgrad}, we show that \gradslam{}  can \emph{fill-in} holes in the depthmap by leveraging multi-view gradient information.}
    \label{fig:e2etasks}
    % \vspace{-0.3cm}
\end{figure*}

\begin{figure*}[!t]
    \centering
    \includegraphics[width=.85\textwidth]{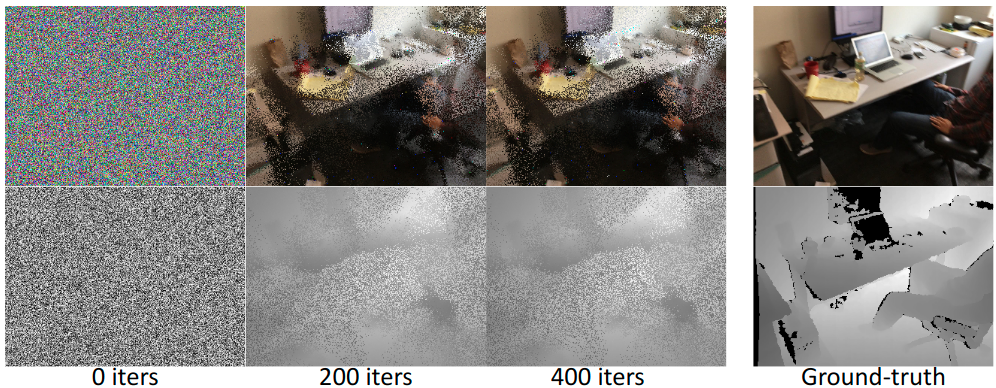}
    \caption{\textbf{RGB-D completion using end-to-end gradient propagation}: Three RGB-D images and a \emph{noise image} are passed through $\nabla$PointFusion, and compared to a clean reconstruction obtained from four RGB-D images. The reconstruction loss is used to optimize the \emph{noise image} by gradient descent. We can recover most of the artifacts from the raw RGB and depth images. Note that finer features are hard to recover from a random initialization, as the overall \emph{SLAM function} is only locally differentiable.}
    \label{fig:rgbdcompletion}
    % \vspace{-0.5cm}
\end{figure*}

In Fig.~\ref{fig:e2etasks}, we similarly introduce such occluders (top row) and pixel noise (bottom row) in one of the depth maps of a sequence and reconstruct the scene using $\nabla$PointFusion. We then calculate the chamfer distance between the noisy and true surfel maps and backpropogate the error with respect to each pixel. The minimized loss leads to the targeted recovery of the noisy and occluded regions. We additionally show an RGB-D image completion task (from uniform noise)in Fig.~\ref{fig:rgbdcompletion}.

% \vspace{-0.25cm}
\section{Conclusion}
\label{sec:conclusion}
% \vspace{-0.1cm}

We introduce \gradslam, a differentiable computational graph framework that enables gradient-based learning for a large set of localization and mapping based tasks, by providing explicit gradients with respect to the input image and depth maps. We demonstrate a diverse set of case studies, and showcase how the gradients propogate throughout the tracking, mapping, and fusion stages. Future efforts will enable \gradslam{} to be directly plugged into and optimized in conjunction with downstream tasks. \gradslam{} can also enable a variety of self-supervised learning applications, as any gradient-based learning architecture can now be equipped with a sense of \emph{spatial understanding}. 

%We believe that a differentiable computation graph framework like \gradslam{} will enable a diverse set of applications. Such frameworks can be directly plugged into and optimized in conjunction with downstream tasks. They also enable a variety of self-supervised learning applications, as any gradient-based learning architecture can now be equipped with a sense of \emph{spatial understanding}.

{\small
\bibliographystyle{ieee_fullname}
\bibliography{references}
}

\appendix

\section{\gradslam{}: Library}

In this paper, we demonstrated that classical dense SLAM systems such as KinectFusion \cite{kinectfusion}, PointFusion \cite{pointfusion}, and ICP-SLAM can all be realized as differentiable computations. However, the set of differentiable modules introduced herein can be used to construct several newer differentiable SLAM systems. To this end, we intend to make the \gradslam{} framework publicly available as open-source software.

\gradslam{} is built on top of PyTorch~\cite{pytorch}, a reverse-mode automatic differentiation library that supports computation over multi-dimensional arrays (often misnomered tensors). At the time of writing this article, \gradslam{} supports the following functionality\footnote{All of these operations are performed fully differentiably.}:

\begin{enumerate}
    \item Non-linear least squares optimization
    \item Depth-based perspective warping (dense visual odometry \cite{kerl2013vo})
    \item Point-to-plane ICP
    \item Raycasting
    \item TSDF volumetric fusion
    \item PointFusion (surfel map building)
    \item ICP-SLAM
    \item Boilerplate operations (Lie algebraic utilities, differentiable vertex and normal map computation, etc.)
\end{enumerate}

\gradslam{} is intended to be an \emph{out-of-the-box} PyTorch-based SLAM framework. Currently, it interfaces with popular datasets such as ScanNet~\cite{scannet}, TUM RGB-D benchmark~\cite{tumrgbd}, and ICL-NUIM~\cite{icl-nuim}.
% By the time of release, we plan on extending functionality to other popular datasets, and also focus on real-time performance.
Currently, low resolution reconstructions (for example, a $128 \times 128 \times 128$ TSDF volume run at about $10 Hz$ on a medium-end laptop GPU (NVIDIA GeForce 1060).

For more details on release timelines, and for more visualizations/results, one can visit \href{http://montrealrobotics.ca/gradSLAM}{this webpage}.

\section{Frequently asked questions (FAQ)}

\begin{enumerate}
    \item \textbf{Q:} \emph{So, \gradslam{} is just classical dense SLAM, implemented using an autograd-compatible language/library?} \\
    \textbf{A:} Yes and no. Technically, while it is possible to ``simply" implement dense SLAM in an autograd-compatible library (eg. PyTorch~\cite{pytorch}), in such a case the obtained gradients would not be meaningful enough, to be used in a gradient-based learning pipeline. We believe that \gradslam{} addresses many such problems (of the gradients being zero ``almost everywhere", akin to impulse functions).
    
    \item \textbf{Q:} \emph{The paper paints a rosy side of \gradslam{}. What are some of the shortcomings of the framework?} \\
    \textbf{A:} Unrolling each computation in dense SLAM as a graph requires an enormous amount of memory. For example, running a differentiable KinectFusion~\cite{kinectfusion} algorithm using a coarse voxel resolution $128 \times 128 \times 128$ ends up requiring $6 GB$ of GPU memory on average. This severely restricts the size of scenes that can be reconstructed in this framework. That is one of the primary concerns we are tackling at the moment. Another aspect of \gradslam{} we are improving upon is to add more robust (differentiable) filters into several stages of the pipeline, such as ICP, photometric warping, etc. We are also working on getting in M-estimators into the optimization routine.
    
    \item \textbf{Q:} \emph{What is the application of such a system?}
    \textbf{A:} We envisage a plethora of applications for a differentiable SLAM system, ranging from enabling spatially-grounded learning, to self-supervision, to task-oriented learning and beyond. We believe that \gradslam{} greatly benefits by following the same modular structure as conventional SLAM systems. This could potentially allow localized learning in only submodules of a SLAM system that actually need to be learnt.
\end{enumerate}

\section{Acknowledgements}

The authors thank Gunshi Gupta for helping out with some of the \gradlm{} experiments. KM would like to thank Ankur Handa for providing valuable feedback/advice on this work. KM also thanks Zeeshan Zia, Ronald Clark, and Sajad Saeedi, who participated in the initial brainstorming sessions. KM and GI acknowledge timely help from Aaditya Saraiya, Parv Parkhiya, Akshit Gandhi, Shubham Garg, all from CMU, who helped capture a live sequence that features in the \gradslam{} video. During the early stages of the project, KM benefitted from participating in brainstorming sessions with Tejas Khot and Gautham Swaminathan. The authors acknowledge the impact of the feedback from colleagues at the Robotics and Embodied AI Lab (REAL), Universit\'e de Montr\'eal on this work; in particular, Gunshi Gupta, Bhairav Mehta, and Mark Van der Merwe.
% KM is grateful to Arun Kumar Singh, for having provided a fix with a ``Type-3 fonts issue", that enabled us to submit this work to a conference, in the nick of time.

\section{Contributions}

\textbf{Krishna Murthy Jatavallabhula} came up with this idea, inspired by an unrelated question on one of his all-nighter coding projects he chose to put up on GitHub \href{https://github.com/krrish94/dvo_python}{https://github.com/krrish94/dvo\_python}. He implemented initial prototypes and played a wrote large parts of the paper.

\textbf{Soroush Saryazdi} led the development of the gradslam library. Although he only joined aboard after our ICRA 2020 submission, he was extremely enthusiastic about the project and was instrumental in its release.

\textbf{Ganesh Iyer} ran several experiments and implemented differentiable ICP and PointFusion pipelines. He led several interesting discussions on follow-up work, and contributed to shaping the core ideas of gradslam.

\textbf{Liam Paull} was the ideal advisor on this project. He supported the rest of us by inspiring us to push harder, and by instilling the spirit of slow science. He wrote a majority of this paper!

\end{document}